\pdfoutput=1
\documentclass[runningheads]{llncs}
\usepackage{booktabs}
\usepackage[table,dvipsnames]{xcolor}
\usepackage{colortbl}
\usepackage{array}
\usepackage{caption}
\usepackage{subcaption}
\usepackage{pgfplots}
\usepackage{pgfplotstable}
\usepackage{tikz}
\usepackage{amsmath}
\usepackage{microtype}
\usepackage{hyperref}
\usepackage{setspace}
\usepackage{amsmath}
\usepackage[T1]{fontenc}
\usepackage{algorithm}
\usepackage{algpseudocode}
\usepackage{tcolorbox}

\usetikzlibrary{shapes.geometric}
\pgfplotsset{compat=1.18}
\usepackage{graphicx}
\begin{document}
\title{Adaptive Arena-based Contestable\\Argumentative Network-of-Experts for Open-Ended Care Plan Coordination}
\titlerunning{Contestable Argumentative Network-of-Experts for Care Coordination}
\authorrunning{T. T. H. Nguyen et al.}
\author{
Truong Thanh Hung Nguyen\inst{1,2,}\thanks{Corresponding author} \and
Hoang-Loc Cao\inst{1} \and
Phuc Ho\inst{1} \and\\
Phuc Truong Loc Nguyen\inst{1} \and
René Richard\inst{2} \and
Hung Cao\inst{1}
}
\institute{
Analytics Everywhere Lab, University of New Brunswick, Canada\\
\and
National Research Council Canada, Canada\\
\email{\{hung.ntt,hung.cao\}@unb.ca, Rene.Richard@nrc-cnrc.gc.ca}
}
\maketitle              
\begin{abstract}
Care plan coordination demands synthesizing heterogeneous clinical, functional, and psychosocial information across multiple professional disciplines, where monolithic LLM pipelines cannot perform in a transparent or safe manner. We present \textit{CANOE (Contestable Argumentative Network-of-Experts)}, a multi-agent neuro-symbolic framework that addresses these limitations through five modules: complexity assessment, adaptive team recruitment, role-based argumentative computation via an \textit{Arena-based Quantitative Bipolar Argumentation Framework (A-QBAF)}, human-in-the-loop contestation, and care-plan synthesis. Role-specialized agents generate supporting and attacking arguments for candidate interventions; conflicts are resolved through arena-based clash resolution before acceptability scores propagate across the argumentation graph. Care planners may accept, reject, edit, or add arguments, and the framework will deterministically recompute the final plan. Evaluation on Discharge Me! and MedicalRAG using ROUGE-L, AlignScore, MEDCON F1, FKGL, and LLM-as-a-judge shows that medically fine-tuned models achieve the strongest clinical correctness and safety, while CANOE's argumentative structure provides faithful explanation and human contestability.
\end{abstract}


\section{Introduction}
Personalized care planning for patients with complex needs, e.g., older adults with multimorbidity, post-discharge patients transitioning home, and individuals living with chronic mental and physical illness, is among the most cognitively and organizationally demanding tasks in modern healthcare. A high-quality care plan must integrate heterogeneous information from electronic health records (EHRs) about a patient's diagnoses, medications, functional limitations, psychosocial circumstances, caregiver capacity and home environment, and reconcile guidance from multiple disciplines (e.g., medicine, nursing, pharmacy, occupational therapy, social work, mental health) under tight time constraints \cite{sinsky2016allocation}. Failures of coordination at this interface are a well-documented driver of avoidable readmissions, medication errors and adverse safety events \cite{ju2022improving}. Reducing this burden has become a central motivation for introducing artificial intelligence (AI) into clinical decision support.

Recent advances in large language models (LLMs) have demonstrated potential for coordinating, summarizing and personalizing care documentation.
LLMs can generate plausible discharge summaries from intensive-care notes, and produce personalized treatment plans when augmented with retrieval and structured prompting \cite{xu2024overview,damm2024wispermed}.
Yet, when deployed for \textit{open-ended} care planning, as opposed to closed multiple-choice question answering, single-model LLM pipelines exhibit several persistent limitations.
First, they hallucinate clinically critical content, e.g., drug dosages, contraindications, and follow-up instructions, with errors that are often plausible enough to evade non-expert review \cite{huang2026hallucinations}. Second, they offer at best post-hoc, narrative chains-of-thought (CoT) \cite{wei2022chain} as explanations; these rationalizations have been shown to be \textit{unfaithful} to the model's actual computations and provide weak guarantees of safety or accountability \cite{freedman_argumentative_2025}. Third, they mimic the perspective of a single generalist clinician, collapsing the multidisciplinary deliberation that characterizes real care planning into one monolithic generation step \cite{liu2026evomdt}. As a result, conflicts between, for example, a pharmacist's concern about polypharmacy and a psychiatrist's preference for a particular antidepressant are silently resolved inside the model, with no mechanism for a human care planner to inspect, edit or contest specific reasoning steps. This directly contradicts emerging regulatory and design principles for high-stakes AI, which demand that systems be contestable-open and responsive to human dispute throughout their lifecycle \cite{nguyen2026heart2mind,ploug_four_2020}.

We argue that progress on AI-assisted care plan coordination requires moving beyond monolithic LLM prompting toward systems that are multidisciplinary by construction, neuro-symbolic in their reasoning, and contestable by design. To this end, we introduce \textit{CANOE (Contestable Argumentative Network-of-Experts)}, a framework for personalized care planning that combines role-based LLM agents with formal argumentative reasoning and structured human-in-the-loop intervention. We evaluate CANOE on BioNLP ACL'24 Discharge Me! \cite{xu2024overview} and MedicalRAG \cite{yao2025control} across three open-source backbones using ROUGE-L, AlignScore \cite{zha2023alignscore}, MEDCON F1 \cite{yim2023aci}, FKGL, and an LLM-as-a-judge protocol scoring Completeness, Validity, Coherence and Safety. The contributions of this paper are as follows:
\begin{itemize}
\item We propose \textit{CANOE (Contestable Argumentative Network-of-Experts)}, the first end-to-end neuro-symbolic framework that decomposes care plan coordination into five modules: complexity assessment and evidence grounding, adaptive multidisciplinary team recruitment, role-based argumentative computation, human-in-the-loop contestation, and care-plan coordination.
\item We introduce an \textit{Arena-based Quantitative Bipolar Argumentation Framework (A-QBAF)} \cite{cao2026neuro} that aggregates supporting and attacking arguments from heterogeneous role-based agents into a deterministically computable acceptability score, providing faithful and auditable rationales for every recommendation.
\item We operationalize contestability by design within a clinical workflow, enabling care planners to accept, reject, edit, or add arguments with formally guaranteed downstream effects on the final plan.
\item We provide an empirical evaluation on two heterogeneous benchmarks with quantitative and qualitative LLM-as-a-judge metrics, offering a reproducible baseline for contestable, multi-agent care plan generation.
\end{itemize}
\section{Related Works}

\subsection{LLM-based Multi-Agent Systems in Healthcare}

A growing body of work argues that role-based collaboration among multiple LLM agents better mirrors real clinical decision making than single-model prompting. MedAgents introduces a multi-disciplinary collaboration framework in which role-playing expert agents engage in multi-round consensus discussion and outperform zero-shot baselines on medical QA \cite{tang2024medagents}. MDAgents extends this idea by adaptively selecting solo, multi-disciplinary or integrated-care-team configurations based on case complexity, achieving state-of-the-art results across ten medical benchmarks \cite{kim2024mdagents}. ClinicalAgent applies multi-agent planning to clinical-trial outcome prediction \cite{yue2024clinicalagent}. While these approaches improve diagnostic accuracy, their dispute resolution typically reduces to majority voting, weighted ensembling or free-form natural-language debate, none of which yields a formally interpretable representation of which arguments drove a final decision.

Computational argumentation offers a principled foundation for this form of structured reasoning. Foundational work established bipolar and defeasible argumentation frameworks as effective tools for medical decision support \cite{xiao2023linked}, and Quantitative Bipolar Argumentation Frameworks (QBAFs) extended these ideas with continuous strength values and gradual semantics that enable the propagation of acceptability scores through attack and support edges \cite{baroni2019fine}. The integration of QBAFs with LLMs is an emerging research direction. ArgLLMs prompt an LLM to generate supporting and attacking arguments and reason over them with gradual semantics, providing faithful explanations and local contestability for binary claim verification \cite{freedman_argumentative_2025}; ArgRAG constructs QBAFs from retrieved documents to combat noisy or contradictory evidence in fact verification \cite{zhu2025argrag}; and ArgMed-Agents brings argumentation schemes into multi-agent clinical reasoning, instantiating an explicit conflict graph and a symbolic solver that selects coherent arguments \cite{hong2024argmed}.

\subsection{Contestable and Explainable AI in Clinical Decision Support}

Explainable AI (XAI) has become a standard requirement for clinical decision support \cite{nguyen2023towards}, motivated by both regulatory pressure (e.g., the EU AI Act and GDPR) and well-documented clinician reluctance to trust opaque models. However, existing approaches remain largely non-interactive, which deliver post-hoc explanations that clinicians cannot interrogate or revise, rather than the narrative, knowledge-grounded justifications that clinical practice demands.

Contestable AI (CAI) has emerged as a complementary paradigm that emphasises human intervention throughout the system lifecycle, encompassing the right to challenge machine predictions, the requirement of human-in-the-loop review for automated decisions, and the need for concrete contestation channels beyond mere explanation \cite{alfrink_contestable_2023,nguyen_motion2meaning_2025}. An AI-supported shared decision-making framework exemplifies this direction in healthcare, calling for AI reasoning to be embedded in clinician-patient deliberation in a transparent and revisable manner \cite{ploug_four_2020,nguyen2026heart2mind}. Recent work further argues that QBAFs can be exchanged between agents to resolve conflicts through argumentative interaction \cite{rago2023interactive,nguyen2026contestable}, and that contestability in automated decision systems requires precisely the kind of structured, interactive reasoning that argumentation frameworks provide \cite{leofante2024contestable}.

CANOE addresses both gaps within a unified system. Unlike existing argumentative LLM systems that target narrow binary tasks \cite{freedman_argumentative_2025,zhu2025argrag,cao2026neuro}, our approach applies an Arena-based extension of QBAFs \cite{cao2026neuro} to multi-disciplinary care plan coordination as an open-ended, multi-objective generation task. To our knowledge, CANOE constitutes the first contestable, multi-agent, neuro-symbolic system designed and evaluated for this setting.

\section{Environment Formulation} 
We consider a decision environment in which the goal is to produce a safe and personalized care plan. This task requires synthesizing clinical conditions, functional abilities, environmental risks, and personal preferences. In practice, high-quality care planning requires contributions from several professional roles, as no single discipline can fully capture the complexity of care plan coordination. Our system aims to support this process by creating a structured multi-agent environment in which different professional viewpoints are revealed, compared, and validated.
Formally, we model the environment as a tuple:
\begin{equation}
    \mathcal{M}=\langle P,\mathcal{D},\mathcal{O},\mathcal{A},\Gamma,H,V,\Pi\rangle.
\end{equation}
Here, the patient information $P$ describes health conditions, functional status, and contextual factors. The system retrieves a set of evidence documents $\mathcal{D}$. A model uses both $P$ and $\mathcal{D}$ to propose a set of candidate care options $\mathcal{O}$. A care team recruitment mechanism then chooses a subset of healthcare roles, denoted $\mathcal{A}$, which serve as agents providing expert analysis. Each agent produces supporting and attacking arguments for each option, creating an argumentative pool $\Gamma$. A human reviewer may revise this set, producing $\Gamma_H$. A validation operator $V$, grounded in quantitative bipolar argumentation semantics, assigns each argument a degree of acceptability. Finally, the care plan coordinator agent $\Pi$ synthesizes a recommended care plan using the weighted argumentative structure.

\section{Framework}\label{sec:implementation}
\begin{figure}[t]
    \includegraphics[width=\linewidth]{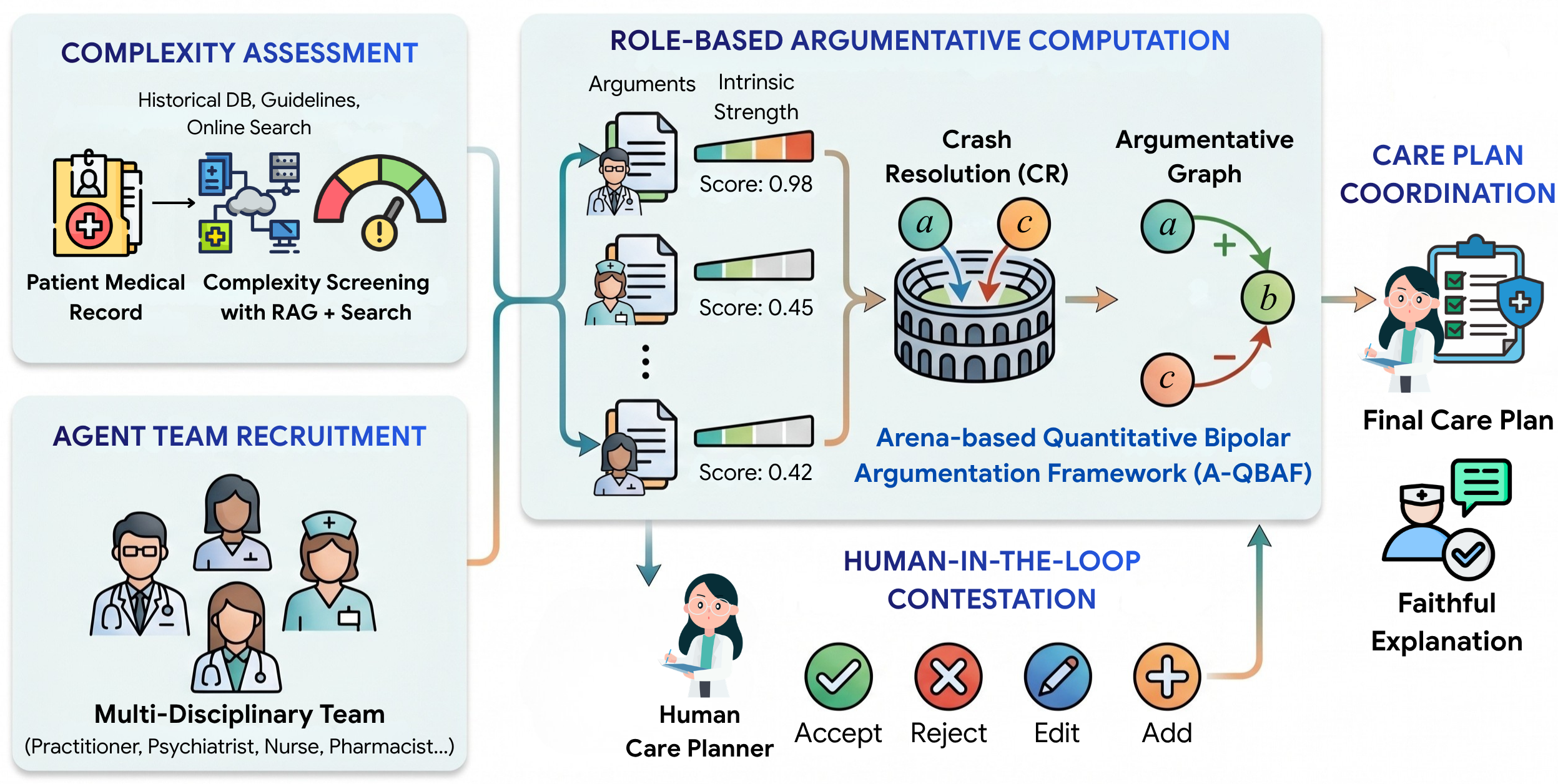} 
    {\caption{Contestable Argumentative Network-of-Experts (CANOE) Framework for Open-Ended Care Plan Coordination.}
    \label{fig:system_overview}}
\end{figure}

In this section, we present \textit{Contestable Argumentative Network-of-Experts} (CANOE), a multi-agent 
neuro-symbolic framework for personalized care planning. The objective of CANOE is to transform 
heterogeneous patient information and retrieved evidence into a transparent, contestable, and 
clinically grounded care plan. Rather than relying on a single model to directly produce 
recommendations, the framework decomposes care planning into five coordinated modules, as shown 
in Fig.~\ref{fig:system_overview} and summarized in Algorithm~\ref{alg:canoe}: (1) complexity 
assessment and evidence grounding, (2) adaptive multidisciplinary team recruitment, (3) role-based 
argumentative computation with an A-QBAF, (4) human-in-the-loop contestation, and (5) care plan 
coordination.

\begin{algorithm}[ht!]
\caption{\small CANOE for Care Planning}
\label{alg:canoe}
\begin{algorithmic}[1]
\small
\Require Patient record $\mathrm{EHR}$, agent pool $\mathcal{A}_{\mathrm{pool}}$, conflict threshold $\delta$, calibration factor $\beta$, human care planner $H$
\Ensure Care plan $Y = \{y_1, \dots, y_q\}$

\vspace{2pt}
\State \textbf{// Module 1: Complexity Assessment \& Evidence Grounding}
\State $P \leftarrow \textsc{ExtractProfile}(\mathrm{EHR})$; \quad $\mathcal{D} \leftarrow \textsc{RetrieveEvidence}(P)$
\State $\mathcal{O} \leftarrow \mathrm{LLM}_{\mathrm{options}}(P, \mathcal{D})$ \Comment{Propose candidate interventions $o_1,\dots,o_m$}
\State $c \leftarrow C(P, \mathcal{D})$, \quad $c \in \{\mathrm{low},\mathrm{moderate},\mathrm{high},\mathrm{very\ high}\}$ \Comment{LLM-based complexity classifier}

\vspace{2pt}
\State \textbf{// Module 2: Adaptive Agent Team Recruitment}
\State $\mathcal{A} \leftarrow S(P, \mathcal{D}, c, \mathcal{O}) \subseteq \mathcal{A}_{\mathrm{pool}}$ \Comment{Broader team if $c \in \{\mathrm{high},\mathrm{very\ high}\}$}

\vspace{2pt}
\State \textbf{// Module 3: Role-Based Argumentative Computation (A-QBAF)}
\State $\Gamma \leftarrow \emptyset$
\For{each $a_j \in \mathcal{A}$, each $o_i \in \mathcal{O}$}
    \State $\Gamma_{j,i} \leftarrow \mathrm{Args}(a_j, o_i, P, \mathcal{D})$ \Comment{$\alpha = [\mathrm{text}(\alpha),\mathrm{stance}(\alpha),\mathrm{role}(\alpha),\mathrm{evidence}(\alpha),\tau(\alpha)]$}
    \State $\tau(\alpha) \leftarrow \mathrm{LLM}_{\mathrm{score}}(\alpha, P, \mathcal{D})$ \quad $\forall\,\alpha \in \Gamma_{j,i}$; \quad $\Gamma \leftarrow \Gamma \cup \Gamma_{j,i}$
\EndFor
\For{each $o_i \in \mathcal{O}$, each opposing pair $(\alpha_s, \alpha_a) \in \Gamma_i$} \Comment{Arena clash resolution}
    \If{$|\tau(\alpha_s) - \tau(\alpha_a)| < \delta$}
        \State $w(\alpha) \mathrel{+}= \mathrm{LLM}_{\mathrm{adj}}(\alpha_s, \alpha_a, P, \mathcal{D})$; \quad $\Delta\tau(\alpha) \leftarrow \beta\bigl(2w(\alpha)-1\bigr)$
        \State $\tilde{\tau}(\alpha) \leftarrow \mathrm{clip}\bigl(\tau(\alpha)+\Delta\tau(\alpha),\;0,\;1\bigr)$
    \EndIf
\EndFor
\For{each $o_i \in \mathcal{O}$} \Comment{A-QBAF propagation}
    \State Construct $\mathcal{Q}_i = \langle X_i, R_i^{+}, R_i^{-}, \tilde{\tau}_i \rangle$; \quad $\tilde{\tau}_i(o_i) \leftarrow 0.5$
    \State Iterate $\sigma_i(x) \leftarrow \tilde{\tau}_i(x) + \bigl(1-\tilde{\tau}_i(x)\bigr)h(E_i(x)) - \tilde{\tau}_i(x)h(-E_i(x))$ until convergence
    \State $F(o_i) \leftarrow \sigma_i(o_i)$
\EndFor

\vspace{2pt}
\State \textbf{// Module 4: Human-in-the-Loop Contestation}
\State Present $\{\mathcal{Q}_i, F(o_i)\}$ to $H$; \quad $\Gamma_H \leftarrow \Gamma$
\For{each action $h_k$ from $H$} \Comment{Accept / Reject / Edit / Add $\alpha \in \Gamma_H$}
    \State Update $\Gamma_H$; \quad construct $\mathcal{Q}_i^{H} = \langle X_i^{H},(R_i^{+})^{H},(R_i^{-})^{H},(\tilde{\tau}_i)^{H}\rangle$
    \State $F_H(o_i) \leftarrow \sigma_i^{H}(o_i)$ \quad $\forall\,o_i \in \mathcal{O}$ \Comment{Deterministic recompute after each edit}
\EndFor

\vspace{2pt}
\State \textbf{// Module 5: Care Plan Coordination}
\State $Y \leftarrow \Pi\!\left(P,\,\mathcal{D},\,\mathcal{O},\,\Gamma_H,\,\{F_H(o_i)\}_{i=1}^{m}\right)$ \Comment{Synthesise plan for $o_i$ with $F_H(o_i)>0.5$}
\State Attach explanation trace (roles, evidence, survived args, scores) to each $y_t \in Y$
\State \Return $Y$
\end{algorithmic}
\end{algorithm}

\subsection{Complexity Assessment and Evidence Grounding}
The framework begins by constructing a structured patient profile from the individual’s demographic data, medical history, current diagnoses, medication regimen, functional limitations, psychosocial conditions, caregiver context, and home-safety factors. We denote this profile by $P=\{p_1,\dots,p_k\}$,
where each $p_\ell$ corresponds to a clinically relevant patient attribute. These attributes are derived primarily from available electronic care records and contextual notes.

To support grounded reasoning, CANOE retrieves evidence from multiple complementary sources. These include historical case records, guideline repositories, and targeted online searches, allowing the framework to combine institution-specific memory with broader clinical knowledge. The retrieved evidence set is denoted by $\mathcal{D}=\{d_1,\dots,d_n\}$.
Each document $d_j$ is associated with metadata (e.g., source type, reliability, retrieval score, and temporal relevance). The purpose is not only to enrich the prompt context but to establish a traceable evidentiary base that can later be linked to specific arguments and recommendations.

Based on $(P,\mathcal{D})$, a planning model proposes a set of candidate care interventions $\mathcal{O}=\{o_1,\dots,o_m\}$, where each option $o_i$ represents a plausible action in the care plan, e.g.,  medication reconciliation, fall-prevention modifications, psychiatric referral, occupational therapy, or community support enrollment.
In parallel, CANOE performs a case-level complexity assessment. Let:
\begin{equation}
    c=C(P,\mathcal{D}) \in \{\textcolor{ForestGreen}{\mathrm{low}}, \textcolor{orange}{\mathrm{moderate}}, \textcolor{red}{\mathrm{high}}, \textcolor{purple}{\mathrm{very\ high}}\},
\end{equation}
denote the estimated complexity level of the care-planning case. The function $C(\cdot)$ summarizes the joint effects of multimorbidity, frailty, cognitive status, polypharmacy, environmental hazards, psychosocial stressors, and care coordination burden. Intuitively, $c$ reflects the degree of interdisciplinary reasoning required to evaluate the candidate interventions safely and comprehensively. Higher-complexity cases, namely those labeled \textcolor{red}{high}/\textcolor{purple}{very high}, require broader expert recruitment and more extensive argumentative analysis, while \textcolor{ForestGreen}{low}/\textcolor{orange}{moderate} cases can be handled by a smaller and more targeted care team.

\subsection{Adaptive Agent Team Recruitment}
CANOE maintains a pool of role-specialized healthcare agents, $\mathcal{A}_{\mathrm{pool}}=\{a_1,\dots,a_r\}$,
where each agent $a$ is defined by a tuple:
\begin{equation}
    a = \langle \mathrm{role}(a), \mathrm{expertise}(a), \mathrm{priorities}(a), \mathrm{style}(a)\rangle.
\end{equation}
Here, $\mathrm{role}(a)$ identifies the healthcare profession (e.g., general practitioner, nurse, pharmacist, psychiatrist, physical therapist, occupational therapist, social worker, nutritionist, care coordinator), $\mathrm{expertise}(a)$ specifies the domains in which the agent is expected to reason reliably, $\mathrm{priorities}(a)$ encodes the clinical dimensions emphasized by that role, and $\mathrm{style}(a)$ governs how the agent formulates recommendations and critiques.

Rather than using the full pool for every patient, the framework adaptively recruits a subset of agents tailored to the current case:
\begin{equation}
    \mathcal{A}=S(P,\mathcal{D},c,\mathcal{O}) \subseteq \mathcal{A}_{\mathrm{pool}}.
\end{equation}

The selection operator $S(\cdot)$ matches the patient profile, retrieved evidences, the candidate intervention set, and the complexity score against the expertise profiles of available roles. For example, a patient with severe depression and medication burden may recruit psychiatry and pharmacy expertise, whereas a patient with frequent near-falls and bathroom hazards may recruit occupational therapy and nursing roles. This recruitment step preserves disciplinary relevance by limiting participation to agents whose perspectives matter for the case. Also, it reduces unnecessary computational overhead while retaining the multidisciplinary structure needed for comprehensive planning.

\subsection{Role-Based Argumentative Computation}
The core reasoning stage of CANOE is a structured argumentative computation process in which each recruited role evaluates the candidate interventions from its own professional perspective. For each option $o_i\in\mathcal{O}$ and each recruited agent $a_j\in\mathcal{A}$, the framework generates role-conditioned arguments grounded in the patient state and retrieved evidence:
\begin{equation}
    \Gamma_{j,i} = \mathrm{Args}(a_j,o_i,P,\mathcal{D}).
\end{equation}

Each generated argument $\alpha \in \Gamma_{j,i}$ is represented as:
\begin{equation}
    \alpha = [\mathrm{text}(\alpha), \mathrm{stance}(\alpha), \mathrm{role}(\alpha), \mathrm{evidence}(\alpha), \tau(\alpha)].
\end{equation}

The field $\mathrm{stance}(\alpha)\in\{\mathrm{support},\mathrm{attack}\}$ indicates whether the argument supports or challenges the adoption of intervention $o_i$; $\mathrm{role}(\alpha)=a_j$ records the professional origin of the argument; $\mathrm{evidence}(\alpha)\subseteq \mathcal{D}$ stores the retrieved sources that substantiate it; and $\tau(\alpha)\in[0,1]$ denotes the argument’s intrinsic strength before considering any interaction with other arguments.

For a fixed intervention $o_i$, the full role-based argumentative pool is $\Gamma_i = \bigcup_{a_j\in\mathcal{A}} \Gamma_{j,i}$, and the global pool across all interventions is $\Gamma = \bigcup_{i=1}^{m}\Gamma_i$.
The intrinsic strength $\tau(\alpha)$ is assigned by a scoring model that evaluates the argument along several dimensions: clinical appropriateness, factual consistency with the retrieved evidence, specificity to the patient’s circumstances, transparency of the reasoning chain, and practical feasibility in the community setting. This scoring stage gives each argument an initial weight, but does not yet account for whether it is reinforced or undermined by other arguments in the discussion.

\subsubsection{Clash Resolution (CR) for Near-Tied Conflicts}
A common failure mode in multi-agent care planning is that a supporting argument and an opposing argument may both appear individually plausible, leading to score saturation and indecisive option ranking. To address this, CANOE implements an arena-based clash-resolution (CR) mechanism before final graph propagation. For any pair of opposing arguments $(\alpha_s,\alpha_a)$ concerning the same intervention $o_i$, where $\alpha_s$ is supportive, and $\alpha_a$ is attacking, the framework checks whether their intrinsic scores are sufficiently close:
\begin{equation}
    |\tau(\alpha_s)-\tau(\alpha_a)| < \delta,
\end{equation}
where $\delta$ is a conflict-sensitivity threshold. If the difference falls below this threshold, the pair is sent to an adjudication module that compares both arguments in light of the patient context and evidence base. The adjudicator determines which argument is more compelling and aggregates outcomes across all such pairwise clashes.
Let $w(\alpha)\in[0,1]$ denote the empirical win rate of argument $\alpha$ across clashes in which it participates. CANOE calibrates intrinsic strengths according to $\Delta\tau(\alpha)=\beta\bigl(2w(\alpha)-1\bigr)$,
where $\beta\geq 0$ controls the magnitude of the calibration. The revised intrinsic score becomes:
\begin{equation}
    \tilde{\tau}(\alpha)=\mathrm{clip}\bigl(\tau(\alpha)+\Delta\tau(\alpha),\,0,\,1\bigr).
\end{equation}

This arena-based calibration sharpens the distinction between near-tied but clinically unequal arguments, ensuring that the final reasoning graph reflects not only raw argument quality but also comparative strength under direct conflict.

\subsubsection{Arena-based Quantitative Bipolar Argumentation Framework}
After generating and calibrating arguments, CANOE organizes them into a formal A-QBAF for each intervention. For option $o_i$, we define:
\begin{equation}
    \mathcal{Q}_i=\langle X_i,R_i^{+},R_i^{-},\tilde{\tau}_i\rangle,
\end{equation}
where $X_i = \{o_i\}\cup\Gamma_i$ is the set of nodes consisting of the intervention node itself and all associated arguments:
\begin{itemize}
    \item $R_i^{+}\subseteq X_i\times X_i$: is the support relation,
    \item $R_i^{-}\subseteq X_i\times X_i$: is the attack relation, 
    \item $\tilde{\tau}_i : X_i \rightarrow [0,1]$: is the calibrated base-strength function.
\end{itemize}

By construction, every supportive argument for $o_i$ yields an edge $(\alpha,o_i)\in R_i^{+}$, and every attacking argument yields an edge $(\alpha,o_i)\in R_i^{-}$. Additional relations may also exist between arguments themselves when one argument semantically reinforces or undermines another. For example, a pharmacist’s warning about sedative burden may support a nurse’s fall-risk concern, while a social worker’s note on caregiver absence may attack the practical feasibility of an otherwise beneficial intervention. In this way, the framework models not only direct supporting/attacking positions on an intervention but also higher-order interactions among the reasons behind those positions.

The intervention node is assigned a neutral prior strength $\tilde{\tau}_i(o_i)=0.5$,
so that its final score emerges from the argumentation process rather than being predetermined. Let $\sigma_i(x)$ denote the propagated acceptability of node $x\in X_i$. For each node, we compute the net argumentative energy:
\begin{equation}
    E_i(x)=\sum_{y\in \mathrm{Sup}_i(x)} \sigma_i(y)-\sum_{y\in \mathrm{Att}_i(x)} \sigma_i(y),
\end{equation}
where $\mathrm{Sup}_i(x)$ and $\mathrm{Att}_i(x)$ are the sets of its supporters and attackers, respectively. Following continuous quantitative bipolar semantics, the propagated score is defined by:
\begin{equation}
    \sigma_i(x)=\tilde{\tau}_i(x) + \bigl(1-\tilde{\tau}_i(x)\bigr)h(E_i(x)) - \tilde{\tau}_i(x)h(-E_i(x)),
\end{equation}
with impact function:
\begin{equation}
    h(z)=\frac{\max\{z,0\}^2}{1+\max\{z,0\}^2}.
\end{equation}

Starting from the calibrated intrinsic scores, the system iterates this update until convergence, yielding stable propagated strengths for every argument and for the intervention node itself.

The option-level acceptability score is then given by the propagated strength of the intervention node $F(o_i)=\sigma_i(o_i)$.
Thus, $F(o_i)$ summarizes how strongly intervention $o_i$ is supported after accounting for role diversity, evidence grounding, inter-argument structure, and direct conflict calibration. Unlike a simple voting or averaging mechanism, this score reflects a genuinely structured deliberation over why an intervention should or should not be included in the care plan.

\subsection{Human-in-the-Loop Contestation}
Although A-QBAF provides a formal and explainable argumentative structure, CANOE is designed as a contestable system rather than a fully autonomous recommender. After the initial role-based computation, the framework presents the reasoning state to the human care team in an editable form. The interface summarizes participation by profession, shows the supporting and attacking arguments for each intervention, links arguments to their evidence sources, and displays both intrinsic and propagated strengths. This makes the rationale for each recommendation inspectable at the level of individual claims and role contributions.
For each argument $\alpha\in\Gamma$, human care planners may perform four classes of actions:
\begin{enumerate}
    \item \textbf{Accept}: retain an argument judged clinically appropriate and well grounded;
    \item \textbf{Reject}: remove an argument deemed unsafe, irrelevant, redundant, or poorly supported;
    \item \textbf{Edit}: revise an argument’s wording, evidence linkage, or relation to other arguments to better reflect the actual patient situation;
    \item \textbf{Add}: introduce a new argument that was missed by the agent team, for example, a contextual consideration arising from caregiver knowledge or local service constraints.
\end{enumerate}
These interventions yield an updated argumentation structure:
\begin{equation}
    \mathcal{Q}_i^{H}=\langle X_i^{H},(R_i^{+})^{H},(R_i^{-})^{H},(\tilde{\tau}_i)^{H}\rangle,
\end{equation}
for each intervention $o_i$. The same propagation procedure is then re-applied to obtain revised acceptability scores $F_H(o_i)=\sigma_i^{H}(o_i)$.

This recomputation is critical as human feedback is not treated as a superficial annotation layer, but as a direct modification of the formal reasoning object. Any accepted change propagates through the support and attack structure and can alter the final ranking of interventions.
In addition, CANOE allows unresolved or borderline interventions to be escalated to the human care planner for final adjudication. This is particularly important in high-risk cases where the propagated score remains near neutral, where competing arguments are clinically consequential, or where feasibility depends on non-documented contextual knowledge. Therefore, the human care planner serves as the final authority, ensuring accountability remains with qualified professionals.

\subsection{Care-Plan Coordination and Faithful Explanation}
The final stage synthesizes the validated intervention set into a coherent care plan. Using the revised option scores $\{F_H(o_i)\}_{i=1}^{m}$, the care-plan coordination module prioritizes interventions, removes incompatible or low-confidence options, and resolves cross-option dependencies such as sequencing, feasibility, and service availability. The output is an integrated care plan that organizes recommendations into a clinically actionable structure, e.g., immediate safety actions, medication-related interventions, mental health follow-up, rehabilitation strategies, caregiver supports, and community service referrals.
Formally, the final care plan is generated by a synthesis care planner:
\begin{equation}
    \Pi(P,\mathcal{D},\mathcal{O},\Gamma_H,\{F_H(o_i)\}_{i=1}^{m}),
\end{equation}
which produces a plan $Y = \{y_1,\dots,y_q\}$,
where each item $y_t$ is a validated recommendation with associated priority, rationale, and evidence trace. Each recommendation is accompanied by a faithful explanation that identifies: (i) which roles supported or challenged it, (ii) which evidence sources were used, (iii) which arguments survived human contestation, and (iv) why the final score justified its inclusion or exclusion. Consequently, CANOE produces care plans that are not only personalized and evidence-based, but also auditable, revisable, and suitable for high-stakes collaborative decision-making.

Overall, CANOE formulates care planning as a contestable process of multidisciplinary argument construction, conflict calibration, formal propagation, and human validation. By integrating adaptive team selection with A-QBAF-based reasoning, the framework converts distributed professional viewpoints into a structured decision object that can be inspected and revised before action is taken. This makes the resulting care plan both clinically meaningful and methodologically transparent.

\section{Experiment and Results}

\subsection{Datasets}
We evaluate care-plan generation using two datasets, i.e., Discharge Me! \cite{xu2024overview} and MedicalRAG \cite{yao2025control}. 
The task is defined as follows: given patient information, including symptoms, diagnoses, clinical history, investigations, hospital course context, and optionally retrieved evidence, the model must generate a clinically appropriate care plan. 
The target care plan is represented as an open-ended, structured plan that includes active problems, recommended tests or monitoring, treatments or medications, follow-up, patient instructions, and return precautions.

\paragraph{\textbf{Discharge Me!}} \cite{xu2024overview} The dataset's original task is to generate the Brief Hospitaßl Course and Discharge Instructions sections of a discharge summary. We adapt this task to care plan generation by treating all non-target patient information as the input and the two target sections as the ground-truth care plan. Specifically, for each admission, we construct the input from the emergency department chief complaint, diagnosis codes, radiology reports, medications or treatments when available, and the discharge summary after removing the Brief Hospital Course and Discharge Instructions sections.

\paragraph{\textbf{MedicalRAG}} \cite{yao2025control} The dataset contains role-specific records, including medical case documents with demographics, history, examination findings, investigations, diagnoses, and plan-like sections such as treatment opinions. We adapt MedicalRAG by filtering to the clinical role \textit{``Medical Practitioners''} and the case type \textit{``records''}. When explicit question–answer fields are available, we retain plan-oriented questions such as ``What further tests are required?'', ``What treatment is recommended?'', or ``What follow-up is needed?'' and use the provided answer as the reference. When only document-style case records are available, we split the document at plan-like headings (i.e., Treatment opinions, Handling opinions, Plan, or Recommendations), whereas the text before this heading is used as the patient input, and the plan-like section is used as the ground-truth output.

\subsection{Experiment Setup}
We evaluate our proposed CANOE performance with three open-weight LLMs that represent distinct points on the capability--specialization trade-off as backbones, i.e., Gemma 4  (8B), MedGemma 1.5 (4B) \cite{sellergren2026medgemma}, and GPT-OSS (20B) \cite{agarwal2025gpt}.
All models are evaluated in a zero-shot prompted setting. All experiments are conducted on a \textit{NVIDIA DGX Spark} device.

\subsection{Metrics}
We evaluate generated care plans using four metrics, capturing four aspects of care-plan quality: ROUGE-L for text similarity, AlignScore for factual grounding in the source evidence, MEDCON for coverage of clinical concepts, and FKGL for readability.
Let $Y_i$ denote the reference care plan, $\hat{Y}_i$ the generated care plan, and $E_i = P_i \cup \mathcal{D}_i$ the combined patient information and retrieved evidence. All metrics are computed at the example level and averaged over the test set $\mathcal{T}$.

\paragraph{\textbf{ROUGE-L}} (Text Similarity) ($\uparrow$) measures textual similarity between $\hat{Y}_i$ and $Y_i$ via the longest common subsequence $\mathrm{LCS}(Y_i,\hat{Y}_i)$. Recall, precision, and F-score:
\begin{equation}
    R_{\mathrm{LCS}} = \frac{\mathrm{LCS}(Y_i,\hat{Y}_i)}{|Y_i|}, \quad
    P_{\mathrm{LCS}} = \frac{\mathrm{LCS}(Y_i,\hat{Y}_i)}{|\hat{Y}_i|},
\end{equation}
\begin{equation}
    \mathrm{ROUGE\text{-}L}(Y_i,\hat{Y}_i) =
    \frac{(1+\beta^2)\,R_{\mathrm{LCS}}\,P_{\mathrm{LCS}}}{R_{\mathrm{LCS}}+\beta^2 P_{\mathrm{LCS}}},
\end{equation}
averaged over $\mathcal{T}$ as $\mathrm{ROUGE\text{-}L}_{\mathcal{T}} = \frac{1}{|\mathcal{T}|}\sum_{i\in\mathcal{T}}\mathrm{ROUGE\text{-}L}(Y_i,\hat{Y}_i)$. Because clinically valid plans may use different wording, ROUGE-L serves as a surface-level supporting metric rather than a direct measure of clinical correctness.

\paragraph{\textbf{AlignScore}} (Factual Consistency) ($\uparrow$) \cite{zha2023alignscore} measures whether the claims in $\hat{Y}_i$ are factually supported by the evidence. Decomposing $\hat{Y}_i$ into atomic claims $\{q_{i1},\ldots,q_{in_i}\}$ and letting $A(q_{ij},\cdot)$ denote a support score, source-grounded scores:
\begin{gather}
    \mathrm{AlignScore}_{\mathrm{src}}(\hat{Y}_i,E_i) = \frac{1}{n_i}\sum_{j=1}^{n_i}A(q_{ij},E_i),
\end{gather}
with the test-set score $\mathrm{AlignScore}_{\mathcal{T}} = \frac{1}{|\mathcal{T}|}\sum_{i\in\mathcal{T}}\mathrm{AlignScore}_{\mathrm{src}}(\hat{Y}_i,E_i)$. This penalizes models that produce plausible but unsupported recommendations.

\paragraph{\textbf{MEDCON}} (Clinical Correctness) ($\uparrow$) \cite{yim2023aci} evaluates clinical correctness by comparing normalized medical concepts. Let $\mathcal{C}(Y_i)$ and $\mathcal{C}(\hat{Y}_i)$ denote concept sets extracted from the reference and generated plans. Precision, recall, and F1 are:
\begin{equation}
    P_{\mathrm{MC}} = \frac{|\mathcal{C}(\hat{Y}_i)\cap\mathcal{C}(Y_i)|}{|\mathcal{C}(\hat{Y}_i)|}, 
    R_{\mathrm{MC}} = \frac{|\mathcal{C}(\hat{Y}_i)\cap\mathcal{C}(Y_i)|}{|\mathcal{C}(Y_i)|},
    F1_{\mathrm{MC}} = \frac{2\,P_{\mathrm{MC}}\,R_{\mathrm{MC}}}{P_{\mathrm{MC}}+R_{\mathrm{MC}}},
\end{equation}
averaged as $F1_{\mathrm{MC},\mathcal{T}} = \frac{1}{|\mathcal{T}|}\sum_{i\in\mathcal{T}}F1_{\mathrm{MC}}(Y_i,\hat{Y}_i)$. Precision measures avoidance of incorrect concepts; recall measures coverage of clinically important ones.

\paragraph{\textbf{Flesch--Kincaid Grade Level (FKGL)}} (Readability) ($\downarrow$) measures readability of the generated plan. Lower values indicate easier readability. Let $W_i$, $S_i$, and $B_i$ denote the number of words, sentences, and syllables in $\hat{Y}_i$:
\begin{equation}
    \mathrm{FKGL}(\hat{Y}_i) = 0.39\!\left(\frac{W_i}{S_i}\right)+11.8\!\left(\frac{B_i}{W_i}\right)-15.59,
\end{equation}
averaged as $\mathrm{FKGL}_{\mathcal{T}} = \frac{1}{|\mathcal{T}|}\sum_{i\in\mathcal{T}}\mathrm{FKGL}(\hat{Y}_i)$.

\subsection{Quantitative Results}
\begin{table}[t!]
\centering
\caption{CANOE evaluation results on Discharge Me! and MedicalRAG. $\uparrow$ higher is better; $\downarrow$ lower is better. Bold = best per column per dataset.}
\label{tab:combined}
\small
\setlength{\tabcolsep}{5pt}
\renewcommand{\arraystretch}{1.2}
\begin{tabular}{l|cccc}
\toprule
\textbf{Model} & \textbf{ROUGE-L}$\uparrow$ & \textbf{AlignScore}$\uparrow$ & \textbf{MEDCON F1}$\uparrow$ & \textbf{FKGL}$\downarrow$ \\
\midrule
\midrule
\rowcolor{gray!15} \multicolumn{5}{c}{\textit{Discharge Me!}} \\
Gemma 4 E4B (8B) & 0.184 & 0.312 & 0.319 & \textbf{11.4} \\
MedGemma 1.5 (4B)  & 0.248 & \textbf{0.451} & \textbf{0.438} & 14.1 \\
GPT-OSS (20B)      & \textbf{0.314} & 0.403 & 0.401 & 12.8 \\
\midrule
\rowcolor{gray!15} \multicolumn{5}{c}{\textit{MedicalRAG}} \\
Gemma 4 E4B (8B) & 0.162 & 0.391 & 0.354 & \textbf{10.9} \\
MedGemma 1.5 (4B)  & 0.238 & \textbf{0.482} & \textbf{0.459} & 13.7 \\
GPT-OSS (20B)      & \textbf{0.289} & 0.428 & 0.436 & 12.3 \\
\bottomrule
\end{tabular}
\end{table}

Table~\ref{tab:combined} reports the quantitative evaluation results across both benchmarks.

\paragraph{Text Similarity} ROUGE-L scores follow model size on both datasets. GPT-OSS scores highest (0.314 on Discharge Me!; 0.289 on MedicalRAG), followed by MedGemma 1.5 (0.248; 0.238) and Gemma 4 E4B (0.184; 0.162). This ordering reflects the ability of larger models to reproduce the surface phrasing of clinical reference text. 

\paragraph{Factual Consistency} AlignScore shows that MedGemma 1.5 ranks first on both datasets (0.451; 0.482), ahead of GPT-OSS (0.403; 0.428) and Gemma 4 E4B (0.312; 0.391). Medical fine-tuning clearly helps the model stay grounded in source evidence, even when its wording differs from the reference. All three models score higher on AlignScore for MedicalRAG than for Discharge Me!, as MedicalRAG cases are shorter and more focused, making them easier to ground against.

\paragraph{Clinical Correctness} MEDCON F1 follows the same ranking as AlignScore. MedGemma 1.5 leads on both datasets (0.438; 0.459), followed by GPT-OSS (0.401; 0.436) and Gemma 4 E4B (0.319; 0.354). The gap between MedGemma 1.5 and GPT-OSS is larger on Discharge Me! (+0.037) than on MedicalRAG (+0.023), because discharge summaries contain denser clinical vocabulary where domain adaptation is more beneficial. Gemma 4 E4B falls 0.119--0.105 points below MedGemma 1.5 on MEDCON F1, confirming that a general-purpose model of this size does not cover enough medical terminology to produce complete care plans.

\paragraph{Readability} Gemma 4 E4B produces the simplest text on both datasets (FKGL 11.4; 10.9), while MedGemma 1.5 produces the most complex (14.1; 13.7), a direct consequence of its exposure to technical biomedical text during fine-tuning. GPT-OSS falls between the two (12.8; 12.3). Since our framework is designed for health professionals, the FKGL range observed across all models is within an acceptable and expected band.

\subsection{LLM-as-a-Judge Evaluation}
\newcounter{template}
\newcommand{\templateref}[1]{Template~\ref{#1}}
\refstepcounter{template}
\label{lst:judge-prompt}
\begin{tcolorbox}[
  title=Template \thetemplate: LLM-as-a-Judge Prompt Template,
  fonttitle=\bfseries\footnotesize,
  colback=gray!5,
  colframe=gray!25,
  coltitle=black
]
\footnotesize
\textbf{System:} You are an expert clinical evaluator. Score the
generated care plan strictly and return only valid JSON.

\textbf{User:}
\texttt{[Patient information], [Retrieved evidence], [Reference plan], [Generated plan]}\\

Score on four dimensions (1--5 Likert scale):
\\
• \textbf{Completeness} -- all clinically relevant sections present
\\
• \textbf{Validity} -- recommendations are evidence-grounded
\\
• \textbf{Coherence} -- logical organization, no contradictions
\\
• \textbf{Safety} -- no harmful or hallucinated content\\

Return JSON: \texttt{\{"completeness": \{"score": <1--5>,
"rationale": "..."\}, ..., "average": <float>\}}
\vspace{4pt}
{\color{gray}\hrule}
\vspace{4pt}
{\scriptsize*Abbreviated for space; full scoring rubrics will be released with the code upon acceptance.
}
\end{tcolorbox}

To complement the quantitative metrics, we evaluate generated care plans using Claude Opus 4.7 as an independent judge, using the structured prompt shown in \templateref{lst:judge-prompt}. For each example, the judge rates the generated plan on four clinician-relevant dimensions, i.e., completeness, validity, coherence, and safety, using a 1--5 Likert scale. This evaluation captures holistic qualities that surface-level metrics such as ROUGE-L and MEDCON F1 cannot fully measure, including logical organization, the presence of unsafe or contradictory recommendations, and whether the plan addresses all clinically relevant aspects of the patient case. Results are
reported in Table~\ref{tab:llm_judge}.

\begin{table}[tb!]
\centering
\caption{LLM-as-a-judge evaluation. Scores are on a 1--5 Likert scale. Bold = best per column per dataset. The average is the mean across dimensions.}
\label{tab:llm_judge}
\small
\setlength{\tabcolsep}{5pt}
\renewcommand{\arraystretch}{1.2}
\begin{tabular}{l|cccc|c}
\toprule
\textbf{Model}
  & \textbf{Completeness}
  & \textbf{Validity}
  & \textbf{Coherence}
  & \textbf{Safety}
  & \textbf{Avg.} \\
\midrule
\midrule
\rowcolor{gray!15}
\multicolumn{6}{c}{\textit{Discharge Me!}} \\
Gemma 4 E4B (8B) & 2.8 & 2.6 & 3.1 & 2.9 & 2.85 \\
MedGemma 1.5 (4B)  & 3.6 & \textbf{4.1} & 3.7 & \textbf{4.0} & \textbf{3.85} \\
GPT-OSS (20B)      & \textbf{3.9} & 3.5 & \textbf{4.2} & 3.4 & 3.75 \\
\midrule
\rowcolor{gray!15}
\multicolumn{6}{c}{\textit{MedicalRAG}} \\
Gemma 4 E4B (8B) & 2.7 & 2.8 & 3.2 & 3.0 & 2.93 \\
MedGemma 1.5 (4B)  & 3.5 & \textbf{4.2} & 3.8 & \textbf{4.1} & \textbf{3.90} \\
GPT-OSS (20B)      & \textbf{3.8} & 3.6 & \textbf{4.3} & 3.5 & 3.80 \\
\bottomrule
\end{tabular}
\end{table}

The judge scores are largely consistent with the quantitative metric findings. 
MedGemma 1.5 achieves the highest average score on both datasets (3.85 on Discharge Me!; 3.90 on MedicalRAG), driven primarily by strong performance on clinical appropriateness (4.1; 4.2) and safety (4.0; 4.1).  
GPT-OSS ranks second overall (3.75; 3.80), with the highest scores on completeness (3.9; 3.8) and coherence (4.2; 4.3). However, its safety scores (3.4; 3.5) are notably lower than those of MedGemma 1.5, suggesting that larger general-purpose models are more prone to including recommendations that are plausible in phrasing but insufficiently grounded in the patient context.
Gemma 4 E4B scores lowest across all dimensions on both datasets, with average scores of 2.85 and 2.93. Its clinical appropriateness scores (2.6; 2.8) are particularly low, consistent with its weak AlignScore and MEDCON F1 results, and confirm that this model does not yet reach a level of clinical reliability suitable for unsupervised care-plan generation.

Altogether, the quantitative and qualitative evaluations point to the same conclusion. MedGemma 1.5 is the strongest model for care-plan generation across both datasets, with consistent advantages on the dimensions most relevant to clinical safety and correctness. GPT-OSS is the most coherent and complete in its outputs, but trades off safety and clinical grounding to achieve this.
\section{Conclusion}
This paper introduced CANOE, a contestable multi-agent neuro-symbolic framework for personalized care plan coordination. Experiments on Discharge Me! and MedicalRAG show that medically fine-tuned models provide the strongest factual consistency and safety, while larger general-purpose models offer greater coherence but weaker clinical grounding. CANOE’s A-QBAF layer makes evidence, conflicts, and care planner interventions explicit and traceable, supporting contestable and safety-sensitive clinical AI. Future work will improve role-specific fine-tuning, argument-strength calibration, and prospective evaluation with practicing care coordinators.
\section{Acknowledgment}
\label{sec:ack}
This work is supported by NSERC Discovery Grant No RGPIN-2025-04478 and NSERC Discovery Supplement Award No DGECR-2025-00129. It is also funded by National Research Council Canada Aging in Place Challenge grant AiP-301-1 D-CGA@home.

\bibliographystyle{splncs04}
\bibliography{references}

@article{agarwal2025gpt,
  author = {Agarwal, Sandhini and others},
  journal = {arXiv preprint arXiv:2508.10925},
  title = {gpt-oss-120b \& gpt-oss-20b model card},
  year = {2025}
}

@article{alfrink_contestable_2023,
  author = {Alfrink, Kars and Keller, Ianus and Kortuem, Gerd and Doorn, Neelke},
  journal = {Minds and Machines},
  note = {Publisher: Springer},
  number = {4},
  pages = {613--639},
  title = {Contestable {AI} by design: {Towards} a framework},
  volume = {33},
  year = {2023}
}

@article{baroni2019fine,
  author = {Baroni, Pietro and others},
  journal = {International Journal of Approximate Reasoning},
  publisher = {Elsevier},
  title = {From fine-grained properties to broad principles for gradual argumentation: A principled spectrum},
  volume = {105},
  year = {2019}
}

@inproceedings{cao2026neuro,
  author = {Cao, Hoang-Loc and others},
  booktitle = {The 39th Canadian Conference on Artificial Intelligence},
  organization = {PMLR},
  pages = {895--902},
  title = {Neuro-Symbolic Adaptive Collaboration of Arena-Based Argumentative LLMs for Contestable Legal Reasoning},
  year = {2026}
}

@inproceedings{damm2024wispermed,
  author = {Damm, Hendrik and others},
  booktitle = {Proceedings of the 23rd Workshop on Biomedical Natural Language Processing},
  pages = {105--121},
  title = {Wispermed at “discharge me!”: Advancing text generation in healthcare with large language models, dynamic expert selection, and priming techniques on mimic-iv},
  year = {2024}
}

@inproceedings{freedman_argumentative_2025,
  author = {Freedman, Gabriel and others},
  booktitle = {Proceedings of the {AAAI} {Conference} on {Artificial} {Intelligence}},
  note = {Issue: 14},
  pages = {14930--14939},
  title = {Argumentative {Large} {Language} {Models} for {Explainable} and {Contestable} {Claim} {Verification}},
  volume = {39},
  year = {2025}
}

@inproceedings{hong2024argmed,
  author = {Hong, Shengxin and others},
  booktitle = {2024 IEEE International Conference on Bioinformatics and Biomedicine (BIBM)},
  organization = {IEEE},
  pages = {5486--5493},
  title = {Argmed-agents: Explainable clinical decision reasoning with llm disscusion via argumentation schemes},
  year = {2024}
}

@inproceedings{huang2026hallucinations,
  author = {Huang, Shengxuan},
  booktitle = {ITM Web of Conferences},
  organization = {EDP Sciences},
  pages = {03005},
  title = {Hallucinations of Large Language Models in Medical Environments: A Systematic Review of Risks, Detection, and Mitigation},
  volume = {84},
  year = {2026}
}

@article{ju2022improving,
  author = {Ju, Hsiao-Hui},
  journal = {The Journal for Nurse Practitioners},
  number = {8},
  pages = {833--836},
  publisher = {Elsevier},
  title = {Improving care coordination of patients with chronic diseases},
  volume = {18},
  year = {2022}
}

@article{kim2024mdagents,
  author = {Kim, Yubin and others},
  journal = {Advances in Neural Information Processing Systems},
  pages = {79410--79452},
  title = {Mdagents: An adaptive collaboration of llms for medical decision-making},
  volume = {37},
  year = {2024}
}

@inproceedings{leofante2024contestable,
  author = {Leofante, Francesco and others},
  booktitle = {Proceedings of the 21st International Conference on Principles of Knowledge Representation and Reasoning},
  pages = {888--896},
  title = {Contestable AI needs computational argumentation},
  year = {2024}
}

@article{liu2026evomdt,
  author = {Liu, Qicai and Hu, Zhichao and Huang, Tao and Niu, Yupeng and Zhang, Xinche and Ma, Shanwu and Lin, Chutong and Huat, Goh Kim and Kwon, Hyeokkoo Eric and Gao, Feng and others},
  journal = {npj Digital Medicine},
  publisher = {Nature Publishing Group UK London},
  title = {EvoMDT: a self-evolving multi-agent system for structured clinical decision-making in multi-cancer},
  year = {2026}
}

@inproceedings{nguyen2023towards,
  author = {Nguyen, Truong Thanh Hung and others},
  booktitle = {International Workshop on Health Intelligence},
  organization = {Springer},
  pages = {11--26},
  title = {Towards trust of explainable ai in thyroid nodule diagnosis},
  year = {2023}
}

@inproceedings{nguyen2026contestable,
  author = {Nguyen, Hung Truong Thanh and others},
  booktitle = {Proceedings of the 2026 International Conference on Multimedia Retrieval},
  pages = {2887--2891},
  title = {Contestable Multi-Agent Debate with Arena-based Argumentative Computation for Multimedia Verification},
  year = {2026}
}

@article{nguyen2026heart2mind,
  address = {New York, NY, USA},
  author = {Nguyen, Hung and Rahimi, Alireza and Whitford, Veronica and Fournier, H\'{e}l\`{e}ne and Kondratova, Irina and Richard, Ren\'{e} and Cao, Hung},
  journal = {ACM Trans. Comput. Healthcare},
  publisher = {Association for Computing Machinery},
  title = {Heart2Mind: Human-Centered Contestable Psychiatric Disorder Prediction System Using Wearable ECG Monitors},
  year = {2026}
}

@inproceedings{nguyen_motion2meaning_2025,
  author = {Nguyen, Loc Phuc Truong and others},
  booktitle = {Proceedings of 9th International Symposium on Chatbots and Human-centred AI},
  title = {{Motion2Meaning}: {A} {Clinician}-{Centered} {Framework} for {Contestable} {LLM} in {Parkinson}’s {Disease} {Gait} {Interpretation}},
  year = {2025}
}

@article{ploug_four_2020,
  author = {Ploug, Thomas and Holm, Søren},
  journal = {Artificial intelligence in medicine},
  title = {The four dimensions of contestable {AI} diagnostics-{A} patient-centric approach to explainable {AI}},
  volume = {107},
  year = {2020}
}

@inproceedings{rago2023interactive,
  author = {Rago, Antonio and Li, Hengzhi and Toni, Francesca},
  booktitle = {Proceedings of the International Conference on Principles of Knowledge Representation and Reasoning},
  number = {1},
  pages = {582--592},
  title = {Interactive Explanations by Conflict Resolution via Argumentative Exchanges},
  volume = {19},
  year = {2023}
}

@article{sellergren2026medgemma,
  author = {Sellergren, Andrew and others},
  journal = {arXiv preprint arXiv:2604.05081},
  title = {Medgemma 1.5 technical report},
  year = {2026}
}

@article{sinsky2016allocation,
  author = {Sinsky, Christine and others},
  journal = {Annals of internal medicine},
  number = {11},
  pages = {753--760},
  publisher = {American College of Physicians},
  title = {Allocation of physician time in ambulatory practice: a time and motion study in 4 specialties},
  volume = {165},
  year = {2016}
}

@inproceedings{tang2024medagents,
  author = {Tang, Xiangru and others},
  booktitle = {Findings of the Association for Computational Linguistics: ACL 2024},
  title = {Medagents: Large language models as collaborators for zero-shot medical reasoning},
  year = {2024}
}

@article{wei2022chain,
  author = {Wei, Jason and others},
  journal = {Advances in neural information processing systems},
  pages = {24824--24837},
  title = {Chain-of-thought prompting elicits reasoning in large language models},
  volume = {35},
  year = {2022}
}

@inproceedings{xiao2023linked,
  author = {Xiao, Liang and Greer, Des},
  booktitle = {Healthcare},
  number = {4},
  organization = {MDPI},
  pages = {585},
  title = {Linked Argumentation Graphs for Multidisciplinary Decision Support},
  volume = {11},
  year = {2023}
}

@inproceedings{xu2024overview,
  author = {Xu, Justin and others},
  booktitle = {Proceedings of the 23rd Workshop on Biomedical Natural Language Processing},
  pages = {85--98},
  title = {Overview of the first shared task on clinical text generation: RRG24 and “discharge me!”},
  year = {2024}
}

@article{yao2025control,
  author = {Hongwei, Yao and others},
  journal = {arXiv preprint arXiv:2504.09593},
  title = {ControlNET: A Firewall for RAG-based LLM System},
  year = {2025}
}

@article{yim2023aci,
  author = {Yim, Wen-wai and others},
  journal = {Scientific data},
  number = {1},
  pages = {586},
  publisher = {Nature Publishing Group UK London},
  title = {Aci-bench: a novel ambient clinical intelligence dataset for benchmarking automatic visit note generation},
  volume = {10},
  year = {2023}
}

@inproceedings{yue2024clinicalagent,
  author = {Yue, Ling and others},
  booktitle = {Proceedings of the 15th ACM International Conference on Bioinformatics, Computational Biology and Health Informatics},
  pages = {1--10},
  title = {Clinicalagent: Clinical trial multi-agent system with large language model-based reasoning},
  year = {2024}
}

@inproceedings{zha2023alignscore,
  author = {Zha, Yuheng and others},
  booktitle = {Proceedings of the 61st Annual Meeting of the Association for Computational Linguistics (Volume 1: Long Papers)},
  pages = {11328--11348},
  title = {AlignScore: Evaluating factual consistency with a unified alignment function},
  year = {2023}
}

@inproceedings{zhu2025argrag,
  author = {Zhu, Yuqicheng and others},
  booktitle = {Conference on Neurosymbolic Learning and Reasoning},
  organization = {PMLR},
  pages = {697--718},
  title = {ArgRAG: Explainable Retrieval Augmented Generation using Quantitative Bipolar Argumentation},
  year = {2025}
}

\end{document}